\documentclass[runningheads]{llncs}
\usepackage{cite}
\usepackage{graphicx}
\usepackage{multirow}
\usepackage{amssymb}
\usepackage{array}
\usepackage{url}
\usepackage{enumerate}
\usepackage{enumitem}
\usepackage{amsmath}
\usepackage{booktabs}
\usepackage{multirow}

\usepackage{algorithm}
\usepackage{algpseudocode}
\usepackage{bm}
\usepackage{bbding}

\usepackage[misc]{ifsym}
\usepackage{lipsum}
\usepackage{color}

\begin{document}

\title{Visual-Attribute Prompt Learning for Progressive Mild Cognitive Impairment Prediction}
\titlerunning{Visual-Attribute Prompt Learning for Progressive MCI Prediction}

\author{
Luoyao Kang\inst{1,2}$^{*}$
\and 
Haifan Gong\inst{1,2}$^{*}$
\and 
Xiang Wan\inst{1}
\and 
Haofeng Li\inst{1}$^{(\textrm{\Letter})}$
}
\authorrunning{L. Kang \textit{et al.}}

\institute{$^1$ Shenzhen Research Institute of Big Data, Shenzhen, China\\ 
$^2$ The Chinese University of Hong Kong, Shenzhen, China\\ 
\email{lhaof@sribd.cn}
}

%
%

\maketitle              

\newcommand\blfootnote[1]{%
\begingroup
\renewcommand\thefootnote{}\footnote{#1}%
\addtocounter{footnote}{-4}%
\endgroup
}

\begin{abstract}
Deep learning (DL) has been used in the automatic diagnosis of Mild Cognitive Impairment (MCI) and Alzheimer’s Disease (AD) with brain imaging data. However, previous methods have not fully exploited the relation between brain image and clinical information that is widely adopted by experts in practice. To exploit the heterogeneous features from imaging and tabular data simultaneously, we propose the Visual-Attribute Prompt Learning-based Transformer (VAP-Former), a transformer-based network that efficiently extracts and fuses the multi-modal features with prompt fine-tuning. Furthermore, we propose a Prompt fine-Tuning (PT) scheme to transfer the knowledge from AD prediction task for progressive MCI (pMCI) diagnosis. In details, we first pre-train the VAP-Former without prompts on the AD diagnosis task and then fine-tune the model on the pMCI detection task with PT, which only needs to optimize a small amount of parameters while keeping the backbone frozen. Next, we propose a novel global prompt token for the visual prompts to provide global guidance to the multi-modal representations. Extensive experiments not only show the superiority of our method compared with the state-of-the-art methods in pMCI prediction but also demonstrate that the global prompt can make the prompt learning process more effective and stable. Interestingly, the proposed prompt learning model even outperforms the fully fine-tuning baseline on transferring the knowledge from AD to pMCI. 
\blfootnote{This work is supported by Chinese Key-Area Research and Development Program of Guangdong Province (2020B0101350001), 
and the National Natural Science Foundation of China (No.62102267), 
and the Guangdong Basic and Applied Basic Research Foundation (2023A1515011464), 
and the Shenzhen Science and Technology Program (JCYJ20220818103001002), 
and the Guangdong Provincial Key Laboratory of Big Data Computing, The Chinese University of Hong Kong, Shenzhen. 

Haofeng Li is the corresponding author (lhaof@sribd.cn). 
Luoyao Kang and Haifan Gong contribute equally to this work.
}

\end{abstract}

\section{Introduction}
Alzheimer’s disease (AD) is one of the most common neurological diseases in elderly people, accounting for 50-70\% of dementia cases~\cite{winblad2016defeating}. 
The progression of AD triggers memory deterioration, impairment of cognition, irreversible loss of neurons, and further genetically complex disorders as well. Mild Cognitive Impairment (MCI), the prodromal stage of AD, has been shown as the optimal stage to be treated to prevent the MCI-to-AD conversion~\cite{spasov2019parameter}. Progressive MCI (pMCI) group denotes those MCI patients who progress to AD within 3 years, while stable MCI (sMCI) patients remained stable over the same time period. pMCI is an important group to study longitudinal changes associated with the development of AD~\cite{risacher2009baseline}. By predicting the pMCI progress of the patients, we can intervene and delay the progress of AD. Thus, it is valuable to distinguish patients with pMCI from those with sMCI in the early stage with a computer-aided diagnosis system. 

In recent years, deep learning (DL) based methods~\cite{arbabshirani2017single,ebrahimighahnavieh2020deep, gong2022vqamix,huang2022attentive} have been widely used to identify pMCI or AD based on brain MRI data. Several works~\cite{el2020multimodal, spasov2019parameter,polsterl2021combining} take both brain MRI and clinical tabular data into account, using convolutional neural networks (CNNs) and multi-layer perceptions as the feature encoder. Due to the limited data in pMCI diagnosis, some works~\cite{lian2018hierarchical,spasov2019parameter} resort to transfer learning to fine-tune the model on the pMCI-related task, by pre-training weights on the AD detection task.  
Previous CNN-based approaches \cite{el2020multimodal, spasov2019parameter,polsterl2021combining} may fail on the lack of modeling long-range relationship, while the models based on two-stage fine-tuning~\cite{lian2018hierarchical,spasov2019parameter} are inefficient and may cause networks to forget learned knowledge~\cite{zhou2022learning}.

Inspired by the advance in transformers\cite{devlin2019bert,li2022view,shaker2022unetr++} and prompt learning\cite{radford2021learning,brown2020language,jia2022visual}, 
we tailor-design an effective multi-modal transformer-based framework based on prompt learning for pMCI detection. The proposed framework is composed of a transformer-based visual encoder, a transformer-based attribute encoder, a multi-modal fusion module, and a prompt learning scheme. Clinical attributes and brain MR images are sent to the attribute encoder and visual encoder with the prompt tokens, respectively. Then the high-level visual and tabular features are aggregated and sent to fully-connected layers for final classification. For the proposed prompt tuning scheme, we first pre-train the neural network on the AD classification task. After that, we fine-tune the neural network by introducing and updating only a few trainable parameters. Importantly, we observe that the number of image patches and input tokens have a gap between 2D natural images and 3D MR images, due to the different dimensions. To complement local interactions between prompt and other tokens, we develop a global prompt token to strengthen the global guidance and make the visual feature extraction process more efficient and stable. 

Our contributions have three folds: (1) We propose a visual-attribute prompt learning framework based on transformers (VAP-Former) for pMCI detection. 
(2) We design a global prompt to adapt to high-dimension MRI data and build a prompt learning framework for transferring knowledge from AD diagnosis to pMCI detection. 
(3) Experiments not only show that our VAP-Former obtains state-of-the-art results on pMCI prediction task by exceeding the full fine-tuning methods, but also verify the global prompt can make the training more efficient and stable. 

\begin{figure*}[!ht]
\centering
\includegraphics[width=\textwidth]{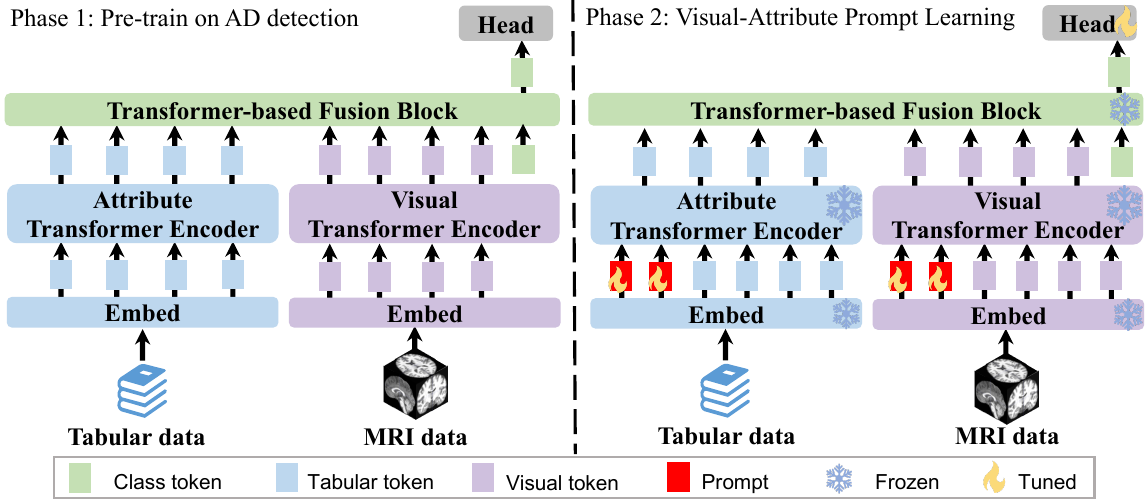}
\caption{The proposed Visual-Attribute Prompt learning transformer framework. The left part shows the architecture of VA-Former without using prompts. The right part shows the proposed framework with the prompt tuning strategy (VAP-Former) that only updates the learnable prompts.} 
\label{fig1}
\end{figure*}

\section{Methodology}
We aim to predict if an MCI patient will remain stable or develop Alzheimer's Disease, and formulate the problem as a binary classification task based on the brain MR image and the corresponding tabular attribute information of an MCI patient from baseline examination. 
As Fig.~\ref{fig1} shows, we adopt the model weights learned from AD identification to initialize the prediction model of MCI conversion. In the prompt fine-tuning stage, we keep all encoders frozen, only optimizing the prompts concatenated with feature tokens.

\subsection{Network Architecture}
We propose a transformer-based framework for pMCI prediction (VAP-Former) based on visual and attribute data, which is shown in Fig.~\ref{fig1}. VAP-Former is mainly composed of three parts: a visual encoder for processing MRI data, an attribute encoder for processing attribute data, and a transformer-based fusion block for combining the multi-modal feature. Considering that capturing the long-range relationship of MRI is important, we employ the encoder of 3D UNETR++\cite{shaker2022unetr++} as our visual encoder. For the attribute encoder, since the clinical variables have no order or position, we embed the tabular data with the transformer blocks~\cite{vaswani2017attention}. 
Followed by \cite{devlin2019bert,dosovitskiy2021an}, we prepend a class token for dual-modality before the transformer-based fusion block. A class token is a learnable vector concatenated with dual-modal feature vectors. 
The class token of dual-modal is further processed by fully-connected layers for the final classification.

\begin{figure*}[!ht]
\centering
\includegraphics[width=\textwidth]{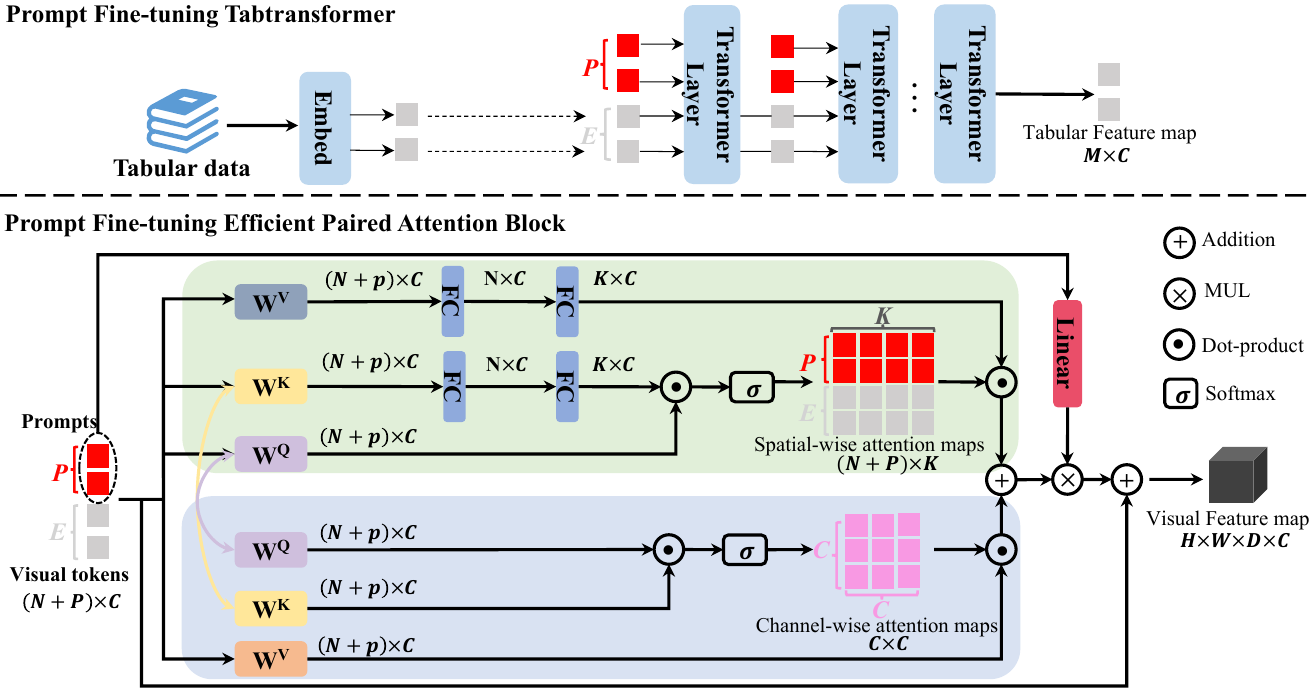}
\caption{The proposed global prompt-based learning strategy. The strategy replaces the attribute encoder with the Prompt Fine-tuning Tab-transformer (upper part) and replaces the Efficient Paired Attention (EPA) block with the prompt fine-tuning EPA block (lower part).} 
\label{fig2}
\end{figure*}

\subsection{Knowledge Transfer with Multi-modal Prompt Learning}
To effectively transfer the knowledge learned from AD prediction task to the pMCI prediction task, we propose a multi-modal prompt fine-tuning strategy that adds a small number of learnable parameters (i.e., prompt tokens) to the input of the transformer layer and keeps the backbone frozen. The overall pipeline is shown in Fig. \ref{fig2}, where the upper part indicates adding the prompt tokens to the attribute transformer, while the lower part denotes sending the prompt tokens to the visual transformer.

\textbf{Tabular Context Prompt Learning.}
In the attribute encoder, we insert prompts into every transformer layer's input~\cite{jia2022visual}. For the $(i)$-th Layer $L_{i}$ of SA transformer block, we denote collection of $p$ prompts as $P = \{p^k\ \in \mathbb{R}^C\|k \in \mathbb{N},1\leq k \leq p\}$, where $k$ is the number of the tabular prompt. The prompt fine-tuning Tabtransformer can be formulated as: 
\begin{equation}
\begin{aligned}
\left[ \underline{\hbox to 4mm{}} , X_i \right] &= L_i([P_{i-1}, X_{i-1}]),\\
\left[ \underline{\hbox to 4mm{}} , X_{i+1} \right] &= L_{i+1}([P_{i}, X_{i}]),
\end{aligned}
\label{eq_tab}
\end{equation}
where $X_i \in \mathbb{R}^{M {\times}C}$ denotes the attribute embedding at $L_i$'s output space and $P_i$ denotes the attribute prompt at $L_{i+1}$'s input space concatenated with $X_i$.

\textbf{Visual Context Prompt Learning with Paired Attention.}
For the visual prompt learning part, we concatenate a small number of prompts with visual embedding to take part in the spatial-wise and channel-wise attention block~\cite{li2019motion,he2019non,li2020depthwise} denoted as the prompt fine-tuning efficient paired attention block in Fig.~\ref{fig2}.
Within the prompt fine-tuning efficient paired attention block, we insert shared prompts into the spatial-wise attention module (SWA) and the channel-wise attention module (CWA), respectively.
With a shared keys-queries scheme, queries, keys, and values are noted as $Q_{shared}^p$, $K_{shared}^p$, $V_{spatial}^p$, and $V_{channel}^p$. Let $[\cdot, \cdot]$ be the concatenation operation, the SWA and CWA can be formulated as:
\begin{equation}
\begin{aligned} 
\left[ P_{spatial},I_S \right] &= SWA(Q_{shared}^p, K_{shared}^p, V_{spatial}^p),\\
\left[ P_{channel},I_C \right] &= CWA(Q_{shared}^p, K_{shared}^p, V_{channel}^p),\\
\end{aligned}
\label{eq_img}
\end{equation}
where $I_S \in \mathbb{R}^{N {\times}C}$ and $P_{spatial}\in \mathbb{R}^{\frac{P}{2} {\times}C}$ are spatial-wise visual embedding and prompts, and $I_C \in \mathbb{R}^{N {\times}C}$ and $P_{channel} \in \mathbb{R}^{\frac{P}{2} {\times}C}$ are channel-wise visual embedding and prompts, and $P$ is the number of visual prompt. After that, the initial feature map $I$ is added to the attention feature map using a skip connection. Let $+$ be the element-wise addition, this process is formulated as $I = I + (I_S+I_C)$.

\subsection{Global Prompt for Better Visual Prompt Learning}

Compared to natural images with relatively low dimensionality (e.g., shaped $224\times 224\times 3$) and the salient region usually locates in a small part of the image, brain MRIs for diagnosis of Alzheimer's are usually high dimensional (e.g., shaped $144\times 144\times 144$) and the cues to diagnosis disease (e.g., cortical atrophy and beta protein deposition) can occupy a large area of the image. Therefore, vanilla prompt learning~\cite{zhou2022learning} methods that are designed for natural images may not be directly and effectively applied to MRI's Recognition of Alzheimer's disease. 
We consider that the above differences lead to the following two problems: (1) vanilla prompt token often focuses on local visual information features; (2) the interaction between vanilla prompt token and visual feature is insufficient. Therefore, we consider that a sophisticated prompt learning module should be able to address the above-mentioned issues with the following feature: (1) the prompt token can influence the global feature; (2) the prompt token can effectively interact with the visual input features. A simple approach to achieve the second goal is to increase the number of prompt tokens so that they can better interact with other input features. However, the experiment proves that this is not feasible. We think it is because too many randomly initialized prompt tokens (i.e., unknown information) will make the model hard to train.
 
Thus, we tailor-design a global prompt token $g$ to achieve the above two goals. Specifically, we apply a linear transformation $T$ to the input prompt tokens $P$ to obtain the global prompt token $g$ (i.e., a vector), which further multiplies the global feature map. Since the vector is directly multiplied with the global feature, we can better find the global feature responses in each layer of the visual network, which enables the model to better focus on some important global features for pMCI diagnosis, such as cortical atrophy. Since this linear transformation, $T$ operation is learnable in the prompt training stage, the original prompt token can better interact with other features through the global token. To embed the global prompt token into our framework, we rewrite Eq.3 as:
\begin{equation}
\begin{aligned}
I &= I + (I_S+I_C) \times T([P_{spatial},P_{channel}]),
\end{aligned}
\label{eq_img3}
\end{equation}
where $T$ denotes the linear transformation layer with $P\times C$ elements as input and $1\times C$ element as output. Experiments not only demonstrate the effectiveness of the above method but also make the training stage more stable.

\section{Experiments and Results}
\subsection{Datasets and Implementation Details}
The datasets are from Alzheimer’s Disease Neuroimaging Initiative (ADNI) \cite{jack2008alzheimer}, including ADNI-1 and ADNI-2. Following \cite{pan2021collaborative,lian2018hierarchical}, we adopt ADNI-1/ADNI-2 as the train/test set. The subjects that exist in both datasets are excluded from ADNI-2. 
There are 1340 baseline T1-weighted structure MRI scans of 4 categories including Alzheimer’s disease (AD), normal control (NC), stable mild cognitive impairment (sMCI), and progressive mild cognitive impairment (pMCI). 
707 subjects (158 AD, 193 NC, 130 pMCI, 226 sMCI) are from ADNI-1, while 633 subjects (137 AD, 159 NC, 78 pMCI, 259 sMCI) are from ADNI-2. 
We pre-process the MR image as \cite{pan2021collaborative}.  All MRI scans are pre-processed via 4 steps: (1) motion correction, (2) intensity normalization, (3) skull stripping, and (4) spatial normalization to a template.
We use the Computational Anatomy Toolbox (CAT12)\protect\footnotemark[1] via Statistical Parametric Mapping software (SPM12)\protect\footnotemark[2] to perform the above procedures. Then all images are re-sampled as the size of 113$\times$137$\times$113 and the resolution of 1$\times$1$\times$1 $mm^{3}$. 
For tabular clinical data, we select 7 attributes including age, gender, education, ApoE4, P-tau181, T-tau, and a summary measure (FDG) derived from 18F-fluorodeoxyglucose PET imaging. 
For tabular clinical data, we apply one-hot encoding to the categorical variables and min-max normalization to the numerical variables. 
\footnotetext[1]{https://neuro-jena.github.io/cat/}
\footnotetext[2]{https://www.fil.ion.ucl.ac.uk/spm/software/spm12/} 
We adopt the model weights learned from AD identification to initialize the prediction model of MCI conversion. Our model is trained using 1 NVIDIA V100 GPU of 32GB via AdamW optimizer \cite{loshchilov2018fixing} with 30 epochs for AD identification and 20 epochs for pMCI detection. The batch size is set to 4. We adopt the ReduceLROnPlateau~\cite{al2022scheduling} learning rate decay strategy with an initial learning rate of 1e-5. The loss function is binary cross-entropy loss. The number of visual and tabular prompts is 10 and 5, respectively. We take F1-score~\cite{pan2020spatially}, class balanced accuracy (BACC)~\cite{brodersen2010balanced}, and the area under the receiver operating characteristic curve (AUC)~\cite{pan2020spatially} as evaluation metrics.

\begin{table}[!ht]
\centering
\caption{Comparison with existing pMCI classification methods. `FT' represents full fine-tuning without prompts and `PT' represents prompt fine-tuning. `\# params' denotes the number of parameters being tuned in the fine-tuning stage.}
\begin{tabular}{cccccccc}
\toprule
    \multirow{2}{*}{Method}    &\multicolumn{2}{c}{Modal}   &\multirow{1}{*}{Fine-}   & \#params  & \multirow{2}{*}{BACC}   &\multirow{2}{*}{F1} &\multirow{2}{*}{AUC}   \\ 
    & Vis& Tab& tuning& (M) &     \\ \midrule
UNETR++~\cite{shaker2022unetr++}&       \multicolumn{1}{|c}{\Checkmark}&     \XSolidBrush&             FT &64.59   & \multicolumn{1}{|c}{$65.94_{\pm1.19}$}     & $47.42_{\pm1.20}$    & $70.46_{\pm1.67}$ \\
Tabformer~\cite{padhi2021tabular}&      \multicolumn{1}{|c}{\XSolidBrush}&     \Checkmark&             FT  &27.13  & \multicolumn{1}{|c}{$78.05_{\pm0.52}$}    & $60.29_{\pm0.27}$    & $84.26_{\pm0.17}$ \\ 
4-Way Classifier~\cite{ruiz20203d}&     \multicolumn{1}{|c}{\Checkmark}&     \XSolidBrush&             FT  &12.27  & \multicolumn{1}{|c}{$71.86_{\pm2.07}$}  & $53.49_{\pm3.44}$    & $75.81_{\pm1.01}$ \\
DFAF~\cite{gao2019dynamic}&             \multicolumn{1}{|c}{\Checkmark}&     \Checkmark&             FT  &69.32  &\multicolumn{1}{|c}{$77.14_{\pm0.35}$}    & $59.66_{\pm0.85}$    & $84.41_{\pm0.49}$ \\ 
HAMT~\cite{chen2021history}&            \multicolumn{1}{|c}{\Checkmark}&     \Checkmark&             FT &71.83  &\multicolumn{1}{|c}{$75.07_{\pm0.91}$}  & $55.58_{\pm1.11}$    & $84.23_{\pm0.31}$ \\ 
DAFT~\cite{polsterl2021combining}&   \multicolumn{1}{|c}{\Checkmark}&     \Checkmark&       FT &67.89  &\multicolumn{1}{|c}{$78.06_{\pm0.55}$}  & $62.92_{\pm0.71}$    & $85.33_{\pm0.41}$ \\ 
\midrule
VA-Former&      \multicolumn{1}{|c}{\Checkmark}&     \Checkmark&          FT &70.19   &\multicolumn{1}{|c}{$78.29_{\pm0.52}$}   & $62.93_{\pm0.29}$    & $84.77_{\pm0.35}$ \\
VAP-Former&    \multicolumn{1}{|c}{\Checkmark}&     \Checkmark&          PT  & 0.59   & \multicolumn{1}{|c}{$\bm{79.22_{\pm0.58}}$}   & $\bm{63.13_{\pm0.11}}$    & $\bm{86.31_{\pm0.25}}$ \\
                                \bottomrule
\end{tabular}
\label{exp1}
\end{table}

\subsection{Comparison with the State-of-the-art Methods}

To validate the proposed VAP-Former and prompt fine-tuning (PT) strategy, we compare our model with three unimodal baselines: 1) UNETR++, which denotes the encoder of UNETR++~\cite{shaker2022unetr++} only using MRI data as input, 2) Tabformer~\cite{padhi2021tabular}, which only uses tabular data and is similar to the attribute encoder work in our model. 3) 4-Way Classifier~\cite{ruiz20203d} which used 3D DenseNet as the backbone to construct Alzheimer's disease diagnosis model only using MRI data as input. To further evaluate the efficiency of our model and the proposed PT strategy, we integrate the DAFT~\cite{polsterl2021combining}, HAMT~\cite{chen2021history}, and DFAF~\cite{gao2019dynamic} into the same attribute and visual encoder for a fair comparison. And we fine-tune the proposed model with two strategies including full fine-tuning (FT) and prompt tuning (PT). 

In Table~\ref{exp1}, the proposed model with PT strategy achieves 79.22\% BACC, 63.13\% F1, and 86.31\% AUC. VAP-Former and VA-Former outperform all unimodal baselines in all metrics, indicating that our model can effectively exploit the relation between MRI data and tabular data to improve the prediction of pMCI. By using the PT strategy, the VAP-Former outperforms the second-best model, DAFT, by 1.16\% BACC, 0.21\% F1, and 0.98\% AUC, demonstrating that the proposed PT strategy can efficiently adapt learned knowledge to pMCI detection and even achieves better classification results. Besides that, VAP-Former achieves the best results by tuning the minimum number of parameters. 

\subsection{Ablation Study and Investigation of Hyper-parameters}
\begin{table}[!th]
\centering 
\caption{Ablation study for prompt fine-tuning. `Vis-GP' denotes the visual global prompt. The best results are in \textbf{bold}.}
\begin{tabular}{ccccccc}
\toprule
    \multirow{2}{*}{Method}  &\multicolumn{2}{c}{Prompt} & \multirow{2}{*}{Vis-GP}& \multirow{2}{*}{BACC} & \multirow{2}{*}{F1} & \multirow{2}{*}{AUC}   \\ 
                            & Visual                        & Tabular        &           &                     &\\ \midrule
VA-Former&    \multicolumn{1}{|c}{\XSolidBrush}&  \XSolidBrush&   \XSolidBrush& \multicolumn{1}{|c}{$78.29_{\pm0.52}$}     & $62.93_{\pm0.29}$    & $84.77_{\pm0.35}$ \\
VisPrompt&        \multicolumn{1}{|c}{\Checkmark}&    \XSolidBrush&   \XSolidBrush&   \multicolumn{1}{|c}{$77.95_{\pm0.66}$}    & $60.15_{\pm2.51}$    & $84.84_{\pm1.28}$ \\
TabPrompt&        \multicolumn{1}{|c}{\XSolidBrush}&  \Checkmark&     \XSolidBrush&   \multicolumn{1}{|c}{$77.39_{\pm0.80}$}    & $60.49_{\pm0.60}$    & $84.70_{\pm0.81}$ \\
Vis-TabPrompt&  \multicolumn{1}{|c}{\Checkmark}&    \Checkmark&     \XSolidBrush& \multicolumn{1}{|c}{$78.19_{\pm0.77}$}    & $63.11_{\pm1.45}$    & $85.23_{\pm0.53}$ \\
VAP-Former&   \multicolumn{1}{|c}{\Checkmark}&    \Checkmark&     \Checkmark&   \multicolumn{1}{|c}{$\bm{79.22_{\pm0.58}}$}  & $\bm{63.13_{\pm0.11}}$    & $\bm{86.31_{\pm0.25}}$ \\
                                \bottomrule
\end{tabular}
\label{exp2}
\end{table}
In this section, we study the internal settings and mechanism of the proposed PT strategy. First, we investigate how the prompts affect the VAP-Former performance. So we remove the visual prompts from the VAP-Former (denoted as TabPrompt in Table~\ref{exp2}) and tabular prompts (denoted as VisPrompt), respectively. Compared with VA-Former, VisPrompt outperforms it by 0.07\% AUC and TabPrompt degrades the performance. However, VAP-Former, which combines tabular prompts and visual prompts, significantly outperforms VA-Former by 1.54\% AUC, indicating that introducing both types of prompts into the model simultaneously results in a more robust pMCI classification model.
We further validate the importance of the global prompt module in the VAP-Former by removing it from the model (denoted as Vis-TabPrompt). As shown in Table~\ref{exp2}, removing the global prompt results in degraded performance by 1.08\% AUC. 

\begin{figure*}[!t]
\centering
\includegraphics[width=0.9\textwidth]{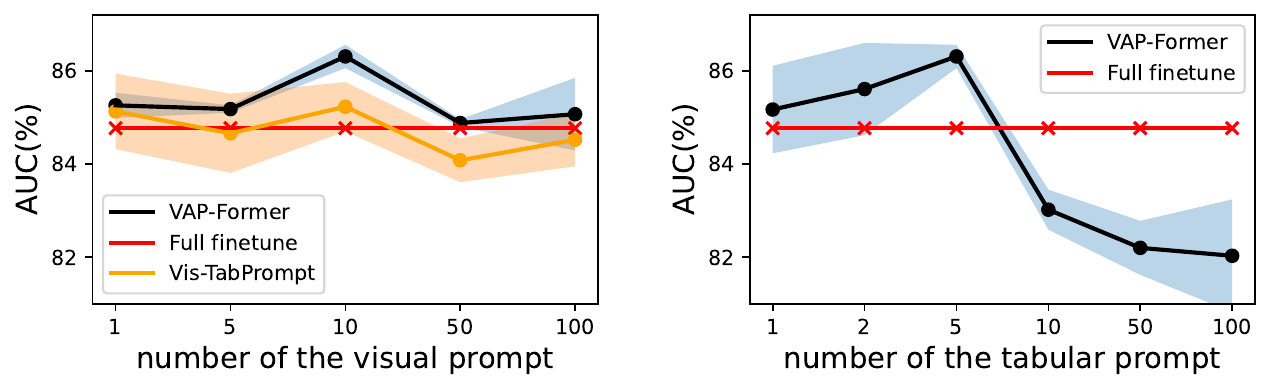}
\caption{Investigation of the number of prompt tokens. The line in red, black, and orange denote the performance of fully-finetuned VA-Former, VAP-Former, and Vis-TabPrompt, respectively. The light blue and orange area represents the error interval obtained by 3 different running seeds. The left figure shows that our global prompt improves the AUC and alleviates the performance fluctuations~\cite{jia2022visual} in prompt learning. 
} 
\label{fig3}
\end{figure*}

To investigate the impact of the number of prompts on performance, we evaluate VAP-Former with varying numbers of prompts. Given that the number of visual tokens exceeds that of tabular tokens. We fix the number of tabular prompts at 5 (left plot in Fig.~\ref{fig3}) and fix the number of visual prompts at 10 (right plot), respectively while changing the other one.
We hypothesize that the interaction between prompts and feature tokens is insufficient, so we gradually increase the number of tokens. As shown in Fig.~\ref{fig3}, we observe that the model's performance ceases to increase after a certain number of prompts, confirming our assumption in Section 3.2 that too many randomly initialized prompt tokens make the model difficult to train.
Conversely, when the number of prompts is small, the prompts can not learn enough information for the task and the interaction between prompt and feature tokens is not sufficient.
Furthermore, as depicted in Fig.~\ref{fig3}, the light blue area is smaller than the orange area, suggesting that the global prompt module makes VAP-Former more robust by helping the model to focus on important global features for pMCI detection.

\section{Conclusion}
To detect pMCI with visual and attribute data, we propose a simple but effective transformer-based model, VA-Former, to learn multi-modal representations. Besides, we propose a global prompt-based tuning strategy, which is integrated with the VA-Former to obtain our overall framework VAP-Former. The proposed framework can efficiently transfer the learned knowledge from AD classification to the pMCI prediction task. The experimental results not only show that the VAP-Former performs better than uni-modal models, but also suggest that the VAP-Former with the proposed prompt tuning strategy even surpasses the full fine-tuning while dramatically reducing the tuned parameters.

\bibliographystyle{splncs04}
\bibliography{paper2041}
\end{document}